\documentclass[lettersize,journal]{IEEEtran}
\usepackage{amsmath,amsfonts}
\usepackage{algorithmic}
\usepackage{algorithm}
\usepackage{array}
\usepackage[caption=false,font=normalsize,labelfont=sf,textfont=sf]{subfig}
\usepackage{textcomp}
\usepackage{stfloats}
\usepackage{url}
\usepackage{verbatim}
\usepackage{graphicx}
\usepackage[nocompress]{cite} 
\usepackage{hyperref}  
\hypersetup{colorlinks}

\usepackage{booktabs}
\usepackage{soul} 
\usepackage{color, xcolor} 
\usepackage{graphicx}
\usepackage{amsmath}
\usepackage{multirow}
\usepackage{colortbl}
\usepackage{bbding}
\usepackage{wrapfig}
\newcommand{\eg}{\textit{e.g.}}

\begin{document}

\title{UEMM-Air: Make Unmanned Aerial Vehicles Perform More Multi-modal Tasks}
% UEMM-Air: A multi-modal and multi-task dataset for unmanned aerial vehicles\textbf{}

\author{Liang Yao, \IEEEmembership{Graduate Student Member, IEEE}, Fan Liu, \IEEEmembership{Member, IEEE}, Shengxiang Xu, Chuanyi Zhang, \IEEEmembership{Member, IEEE}, Xing Ma, Jianyu Jiang, Zequan Wang, Shimin Di, \IEEEmembership{Member, IEEE}, and Jun Zhou, \IEEEmembership{Senior Member, IEEE}
\thanks{Corresponding author: Fan Liu (fanliu@hhu.edu.cn).}
\thanks{Fan Liu, Liang Yao, Shengxiang Xu, Xing Ma, Jianyu Jiang, and Zequan Wang are with the College of Computer Science and Software Engineering, Hohai University, Nanjing, 210098, China.}
\thanks{Chuanyi Zhang is with the College of Artificial Intelligence and Automation, Hohai University, Nanjing, 210098, China.}
\thanks{Shimin Di is with the Department of Computer Science and Engineering, Hong Kong University of Science and Technology, Clear Water Bay, Kowloon Hong Kong.}
\thanks{Jun Zhou is with the School of Information and Communication Technology, Griffith University, Nathan, Queensland 4111, Australia.}% <-this % stops a space
}
%\thanks{Manuscript received April 19, 2021; revised August 16, 2021.}

% The paper headers
\markboth{Journal of \LaTeX\ Class Files,~Vol.~14, No.~8, August~2021}%
{Shell \MakeLowercase{\textit{et al.}}: A Sample Article Using IEEEtran.cls for IEEE Journals}

\IEEEpubid{0000--0000/00\$00.00~\copyright~2021 IEEE}
% Remember, if you use this you must call \IEEEpubidadjcol in the second column for its text to clear the IEEEpubid mark.

\maketitle

\begin{abstract}
The development of multi-modal learning for Unmanned Aerial Vehicles (UAVs) typically relies on a large amount of pixel-aligned multi-modal image data. However, existing datasets face challenges such as limited modalities, high construction costs, and imprecise annotations. To this end, we propose a synthetic multi-modal UAV-based multi-task dataset, \textbf{UEMM-Air}. 
Specifically, we simulate various UAV flight scenarios and object types using the Unreal Engine (UE). 
Then we design the UAV's flight logic to automatically collect data from different scenarios, perspectives, and altitudes. 
Furthermore, we propose a novel heuristic automatic annotation algorithm to generate accurate object detection labels. 
Finally, we utilize labels to generate text descriptions of images to make our UEMM-Air support more cross-modality tasks.
In total, our UEMM-Air consists of \textit{120k} pairs of images with \textit{6} modalities and precise annotations. 
Moreover, we conduct numerous experiments and establish new benchmark results on our dataset. We also found that models pre-trained on UEMM-Air exhibit better performance on downstream tasks compared to other similar datasets. The dataset is publicly available (\url{https://github.com/1e12Leon/UEMM-Air}) to support the research of multi-modal tasks on UAVs. 
\end{abstract}

\begin{IEEEkeywords}
Unmanned Aerial Vehicles, Large Scale Dataset, Multi-modal, Multi-task
\end{IEEEkeywords}

\section{Introduction}
% 无人机飞速发展，有很大潜力。当前大多数任务局限在视觉上。但无人机由于视角问题，通用领域的数据不好直接使用。为支持无人机视觉任务，构建了很多专门的无人机数据集。
With the advancement of Unmanned Aerial Vehicles (UAV) technology~\cite{huang2022object,luo2023evolutionary, khankeshizadeh2024novel,cheng:uav-system,srivastava2023techniques} and deep learning~\cite{lecun2015deep,sharifani2023machine}, vision perception tasks of UAV have shown great potential in many fields such as urban monitoring, military reconnaissance and rescue~\cite{yan2023uav,srivastava2023techniques,su2023ai,liu2022military}.
Unlike general vision task~\cite{zou2023object}, UAV tasks exhibit characteristics such as complex backgrounds, varying scales, and small objects.
Therefore, models trained on general  datasets~\cite{lin2014microsoft,everingham2010pascal} can hardly be directly applied to UAV platforms. To this end, many scholars have constructed vision datasets from the perspective of UAV. 
%For example, the object detection from the perspective of UAV (UAV-OD)  has attracted widespread attention from research community. 
For example, VisDrone~\cite{du2019visdrone} and UAVDT~\cite{du2018unmanned} consider various scenes, weather conditions, and environments, providing good benchmarks for UAV-based Object Detection (UAV-OD)~\cite{mittal2020deep,wu2021deep,zitar2023review,zhang2024empowering} tasks. SkyScenes~\cite{khose2023skyscenes} encompassing different layouts, weather, and daytime conditions with corresponding dense annotations and viewpoint metadata for UAV-based Segmentation (UAV-Seg)~\cite{osco2021review,ahmed2024advancements}. The aforementioned datasets make significant contributions to traditional UAV vision tasks.
% xxx 数据集用于分割任务

\begin{figure}[t]
  \centering
  \includegraphics[width=0.99\linewidth]{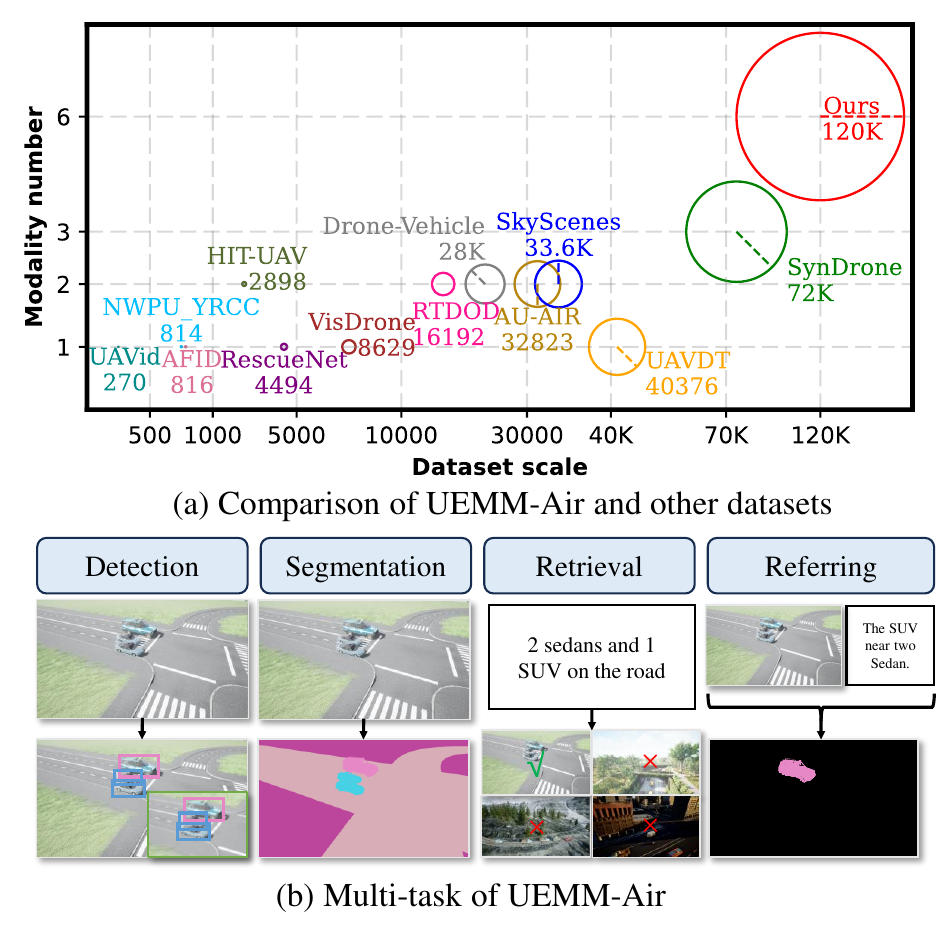}
  %\vspace{-0.15cm}
  \caption{(a) Comparison of UEMM-Air and other UAV environmental perception datasets. (b) Various UEMM-Air targeted tasks. Our stands out as the largest in terms of data scale, featuring the most paired modality types and the greatest variety of tasks among existing datasets.}
  %\vspace{-0.6cm}
  \label{fig1}
\end{figure}

\IEEEpubidadjcol
However, with the development of multi-modal learning~\cite{blikstein2013multimodal,xu2023multimodal}, the above datasets are facing challenges such as being single-modality and having insufficient data. 
Therefore, the research communities are gradually moving from single-modal to multi-modal tasks. For instance,
DroneVehicle~\cite{sun2022drone} utilizes two image modalities: infrared and visible, with the infrared modality enhancing detection accuracy in nighttime scenes.
However, due to the request for manual labeling and aligning two modalities images, the dataset annotation cost is relatively high. %Besides, such type of datasets typically only have two aligned image modalities. 
Although AU-AIR dataset~\cite{bozcan2020air} takes into account the potential value of UAV parameters, it covers relatively only a few scenes and its labeled samples have some imprecise annotations. Drawbacks in above datasets are adverse to model training. 
% These drawbacks are not conducive to model training. 
Furthermore, all the existing datasets are unable to support alignment of vision and language to perform zero-shot inference like the CLIP~\cite{radford2021learning} model in the field of UAVs. Similar to satellite remote sensing, UAVs also need numerous vision-language applications such as open-vocabulary object detection~\cite{wu2024towards}, referring image segmentation~\cite{qiao2020referring}, and multi-modal large language models (MLLMs)~\cite{yin2023survey}, etc.

\begin{figure*}[t]
    \centering
    \includegraphics[width=0.99\linewidth]{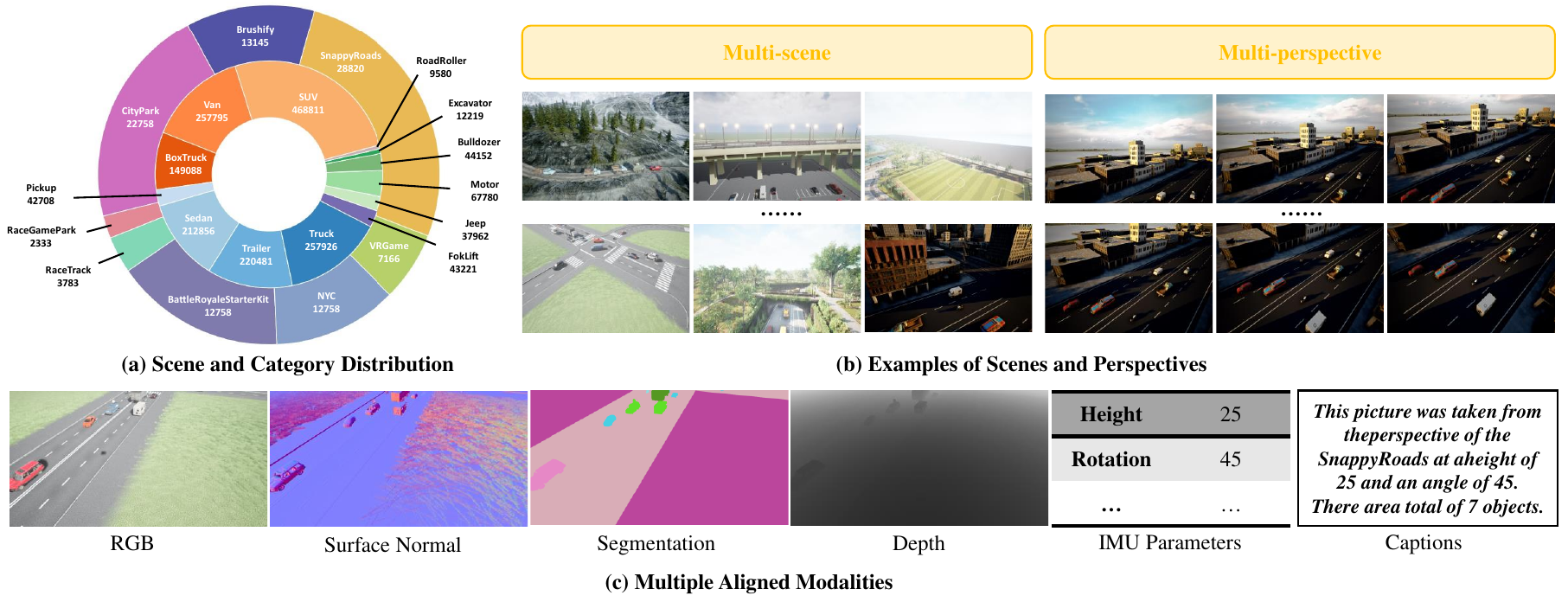}
    \caption{UEMM-Air is a multi-scene, multi-modal, and multi-perspective UAV-based perception dataset. (a) Scence (outer) and object category (inner) distribution. (b) UEMM-Air features multiple scenes and various perspectives of the same view. (c) UEMM-Air encompasses 6 modalities: RGB, surface normals, segmentation, depth, IMU parameters, and textual descriptions.
    }
    \label{fig2}
\end{figure*}

Motivated by the above observations, we construct a new large-scale synthetic UAV environmental perception dataset, UEMM-Air, to facilitate further research on the UAV field. As represented in Fig.~\ref{fig1} and Fig.~\ref{fig2}, we assign UEMM-Air three significant characteristics. \textbf{(1) Multi-modal:} our dataset contains six modalities, including visible, depth, segmentation, surface normals, UAV IMU parameters, and captions. \textbf{(2) Multi-task:} our dataset is capable of supporting tasks such as object detection, instance segmentation, image-text contrastive learning, and referring image segmentation, etc. \textbf{(3) Multi-semantic:} our dataset covers a variety of scenarios and perspective, and has fine-grained category information.

To achieve these objectives, we first utilize the Unreal Engine (UE)~\cite{unrealengine} and AirSim~\cite{airsim2017fsr} framework to build various simulated scenarios for UAV flights. Subsequently, we implement automatic UAV flight control and collect data at different altitudes, scenes, and modalities. 
Furthermore, we design an annotation algorithm to automatically generate object detection labels. 
Finally, we generate text descriptions for different cross-modal tasks according to existing detection and segmentation annotations. 
%Additionally, we conduct experiments on multiple types of tasks in the field of object detection, demonstrating the research significance of our dataset across various tasks.
We also implement representative and impressive methods to systematically investigate the potential and challenges brought by UEMM-Air. The experimental results demonstrate the research significance of our dataset across various tasks.

The main contributions of this article are as follows:

\begin{itemize}
    \item We propose a new synthetic UAV-based environmental perception dataset, UEMM-Air. To the best of our knowledge, \textbf{it is the largest dataset in terms of existing data scale, featuring the most paired modality types and the highest variety of task types.}
    \item We introduce a new heuristic algorithm for automatic data annotation. Compared with labeling strategies in other synthetic datasets, ours can provide more accurate annotations by introducing segmentation and depth modalities to enhance the identification of objects, especially in addressing visually overlapped objects.
    % It can avoid the issue of incorrect annotations caused by generative labeling.  
    \item We conduct experiments on multiple types of tasks in the field of UAV, providing new benchmark results. Our UEMM-Air expands the field of UAV environmental perception from purely visual to multi-modal tasks.
    %Due to our accurate labeling and rich modalities, our UEMM-Air provides remarkable transferability for downstream tasks.
\end{itemize}

The remainder of this paper is structured as follows. In Section~\ref{RW}, we review related literature on UAV environmental perception and existing datasets. Section~\ref{UEMM-Air} introduces our UEMM-Air. Specifically, Section~\ref{fly} introduces the automatic data collection methods we used. Then, in Section~\ref{setup1}, we discuss the configurations and advantages of each modality. We also outline the strategies for automatic annotation generation and the methods for cross-modal generation, respectively in Sections~\ref{auto} and Section~\ref{Cross}.
Next, Section~\ref{experiments} validates the accuracy of our annotations and demonstrates the benchmarks of our UEMM-Air across multiple tasks, including object detection, instance segmentation, referring segmentation, and cross-modal retrieval. Finally, Section~\ref{discussion} presents our discussion of the advantages and limitations of our UEMM-Air and concludes this paper.

\begin{table*}[t]
\caption{Comparison of different UAV-based datasets. `Det': Object Detection, `Seg': Semantic Segmentation, `Ref': Referring Image Segmentation, `CL': Image-Text Contrastive Learning. `MM': Multi-modal. `Angle': UAV's Pan\&Tilt view angle. `-': not applicable or not explicit in their papers.}
\resizebox{\textwidth}{!}{
\begin{tabular}{clccccccc}
\toprule
Tasks    &Dataset    & Year    & MM    & \# modalities    & \# images    & \# classes    & Size [px]    & Angle  \\
\hline
\multicolumn{1}{c|}{\multirow{8}{*}{Det}} &  Stanford-Drone~\cite{Robicquet2016LearningSE}    & 2016    & \XSolidBrush    & -    & -    & 7    & 1450×1080    & 90 \\
\multicolumn{1}{c|}{} & UAVDT~\cite{du2018unmanned}    & 2018     & \XSolidBrush    & -    & 40376    & 3    & 1080×540    & variable \\
\multicolumn{1}{c|}{} & VisDrone~\cite{du2019visdrone}    & 2018    & \XSolidBrush    & -    & 8629    & 10    & 1920×1080    & variable \\
\cline{2-9}
\multicolumn{1}{c|}{} & AU-AIR~\cite{bozcan2020air}    & 2020   & \Checkmark    & 2    & 32823    & 8    & 1920×1080    & 45 to 90 \\
\multicolumn{1}{c|}{} & Drone-Vehicle~\cite{sun2020drone}    & 2022    & \Checkmark    & 2    & 28$k$   & 5    & 640×512    & 90 \\
\multicolumn{1}{c|}{} & HIT-UAV~\cite{suo2023hit}    & 2023    & \Checkmark    & 2    & 2898    & 4    & 640×512    & 30 to 90 \\
\multicolumn{1}{c|}{} & RTDOD~\cite{feng2023rtdod}    & 2023    & \Checkmark    & 2    & 16192   & 10    & 1280×720   & variable \\
\multicolumn{1}{c|}{} & State-Air~\cite{liu2024boost}    & 2024    & \Checkmark    & 2    & 2864   & 4    & 1280×720   & variable \\
\hline
\multicolumn{1}{c|}{\multirow{5}{*}{Seg}} & UAVid~\cite{lyu2020uavid}    & 2020    & \XSolidBrush    & -    & 270   & 8   & 3840×2160   & 45 \\
\multicolumn{1}{c|}{} & NWPU\_YRCC~\cite{zhang2020icenet}    & 2020     & \XSolidBrush    & -    & 814    & 3    & 1600×640    & variable \\
\multicolumn{1}{c|}{} & RescueNet~\cite{rahnemoonfar2023rescuenet}    & 2022     & \XSolidBrush    & -    & 4494    & 10    & 3000×4000    & variable \\
\multicolumn{1}{c|}{} & AFID~\cite{wang2023aerial}    & 2023     & \XSolidBrush    & -    & 816    & 8    & 2560×1440    & variable \\
\multicolumn{1}{c|}{} & SkyScenes~\cite{khose2023skyscenes}    & 2024     & \Checkmark    & 2    & 33.6$k$   & 28    & 2160×1440    & variable \\
\hline
\multicolumn{1}{c|}{Dets, Seg} & SynDrone~\cite{10350525}    & 2023    & \Checkmark    & 3    & 60$k$    & 9    & 1920×1080    & 30,60,90 \\
\hline
\multicolumn{1}{c|}{Dets, Seg, Ref, CL} & \cellcolor[HTML]{DAE8FC}\textbf{UEMM-Air (Ours)}    & \cellcolor[HTML]{DAE8FC}2025    & \cellcolor[HTML]{DAE8FC}\Checkmark    & \cellcolor[HTML]{DAE8FC}\textbf{6}    & \cellcolor[HTML]{DAE8FC}\textbf{120$k$}    & \cellcolor[HTML]{DAE8FC}\textbf{13}    & \cellcolor[HTML]{DAE8FC}1920×1080    & \cellcolor[HTML]{DAE8FC}variable \\
\bottomrule
\end{tabular}
}
%\vspace{-0.3cm}
\label{tab:datasets}
\end{table*}

\section{Related Work} \label{RW}

%这只是一段占位符
\subsection{UAV-based Environmental Perception} 
UAV-based environmental perception primarily involves two tasks: object detection and semantic segmentation.
\subsubsection{Object Detection}
Due to the typically top-down perspective of UAVs and the relatively small scale of the objects, general object detection methods~\cite{RCNN,FasterRCNN,yolo,yolov4,lv2023detrs} tend to be not suitable for UAV-based Object Detection (UAV-OD) tasks. The mainstream methods for UAV-based object detection primarily employ coarse-to-fine strategies~\cite{duan2021coarse,li2020density,yang2019clustered}. Initially, the detection process focuses on identifying larger objects, while concurrently pinpointing dense subregions containing small objects. These subregions are subsequently utilized as inputs for the model to refine detection results. For example, a CZDetector~\cite{CZDetector} employed a density-based cropping algorithm to identify regions with crowded objects and then increased the size of those regions to enhance the training dataset. Alternatively, ~\cite{koyun2022focus} utilizes a Gaussian mixture model to supervise the detector in generating object clusters composed of focusing regions.
To address limited computing resources, methods were proposed to balance accuracy and efficiency. Typical examples include CEASC~\cite{CEASC} and SIFDAL~\cite{liu2024boost}. The former adopted a plug-and-play detection method with enhanced sparse convolution and an adaptive mask scheme. The latter disentangled scale-invariant features to boost detection accuracy and mildly reduce test inference costs. Additionally, to adapt to the low computational power conditions of UAVs, some researchers~\cite{yao2024domain,zou2024remotetrimmer} employed compress techniques such as pruning and distillation.

\subsubsection{Semantic Segmentation}
Similar to object detection, there are many semantic segmentation methods specifically designed for UAV scenarios. High-resolution representation learning plays a crucial role in UAV semantic segmentation due to the ultra-high resolution and varying object scale of UAV remote sensing imagery~\cite{xie2021super,liu2021study,ye2022post}. For example, HRNet~\cite{wang2020deep} was designed by a high-resolution network that repeatedly exchanges semantic information across adjacent multi-resolution sub-networks. At the same time, many lightweight methods were proposed to meet the real-time application requirements of UAV platforms~\cite{orsic2019defense,broni2023real,cheng2024methods}. Literature~\cite{pang2022sgbnet} also proposed Semantics Guided Bottleneck Network (SGBNet) based on BiSeNet and the Channel Pooled Attention (CPA) mechanism to balance segmentation accuracy, model size, and inference speed on the Land Cover Dataset.

\subsection{UAV-based Environmental Perception Datasets} 
% \subsubsection{Object Detection Datasets}
Many UAV-based environmental perception datasets provide multi-class images and videos captured by UAVs. These datasets are significant for promoting the research and development of various computer vision tasks, including object detection, object tracking, and scene understanding. We summarized several commonly used UAV-based datasets in Table~\ref{tab:datasets}. 
%Due to the high cost of obtaining images taken by UAV, early datasets, such as UAVDT and VisDrone, have only a single modality, namely RGB images. In addition, owing to hardware limitations, datasets like Drone-Vehicle are small, low-resolution, and difficult to expand.

\textbf{Stanford-Drone}~\cite{Robicquet2016LearningSE} is a large-scale dataset containing overhead images and videos of multi-class objects moving and interacting at Stanford University. This dataset can be used for learning and evaluating multi-object tracking, activity understanding, and trajectory prediction.

\textbf{UAVDT}~\cite{du2018unmanned} has 80,000 representative frames which are annotated with bounding boxes and 14 kinds of attributes in various complex scenarios. It focuses on 3 specific computer vision tasks: object detection, single-object tracking, and multiple-object detection.

\textbf{VisDrone}~\cite{du2019visdrone} is a large-scale benchmark dataset in object detection and tracking with various environment conditions and camera viewpoints. It contains 10 categories objects of frequent interest in drone applications and more than 2.5 million annotation bounding boxes.

\textbf{AU-AIR}~\cite{bozcan2020air} includes extracted frames meta-data, bounding box annotations for traffic-related object categories, and multi-modal flight sensor data. The dataset is captured at low altitudes at the intersection.  

\textbf{Drone-Vehicle}~\cite{sun2020drone} offers a drone-based RGB-Infrared cross-modality vehicle detection dataset and corresponding precise annotations. This dataset covers multiple scenarios and objects from day to night with three different angles and heights.

\textbf{HIT-UAV}~\cite{suo2023hit} is the first publicly available high-altitude UAV-based infrared thermal dataset for object detection applications on UAVs. The dataset provides two types of bounding box annotations (oriented and standard) to tackle the challenge of overlapping object instances in aerial images.

\textbf{RTDOD}~\cite{feng2023rtdod} is the first RGB-Thermal domain-incremental object detection dataset designed specifically for UAVs. The dataset covers a wide range of weather conditions and day-to-night transitions, highlighting the complexity of real-world scenarios.

\textbf{State-Air}~\cite{liu2024boost} is an aerial dataset with multi-modal sensor data collected in real-world outdoor environments. The dataset concludes 2246 images of sunny days and 616 instances of snowy ones with 4 categories: person, car, bus, and van. 

% \subsubsection{Semantic Segmentation Datasets}
\textbf{UAVid}~\cite{lyu2020uavid} is a semantic segmentation dataset designed for UAV semantic segmentation in complex urban scenes, featuring on both static and moving object recognition. It provides 300 high-resolution oblique-view UAV images, labeled with 8 classes, and gives a diverse representation of objects with rich scene context.

\textbf{NWPU\_YRCC}~\cite{zhang2020icenet} is the first public UAV image dataset containing 814 accurately annotated images for river ice segmentation. It covers typical images of river ice in different periods, with diverse appearances and captured from different flight heights and views.

\textbf{RescueNet}~\cite{rahnemoonfar2023rescuenet} propose a high-resolution post-disaster dataset for natural disaster damage assessment. It includes detailed classification and semantic segmentation annotations, enabling applications in building damage classification, road segmentation, and future disaster assessment.

\begin{figure*}[t]
  \centering
  \includegraphics[width=0.99\linewidth]{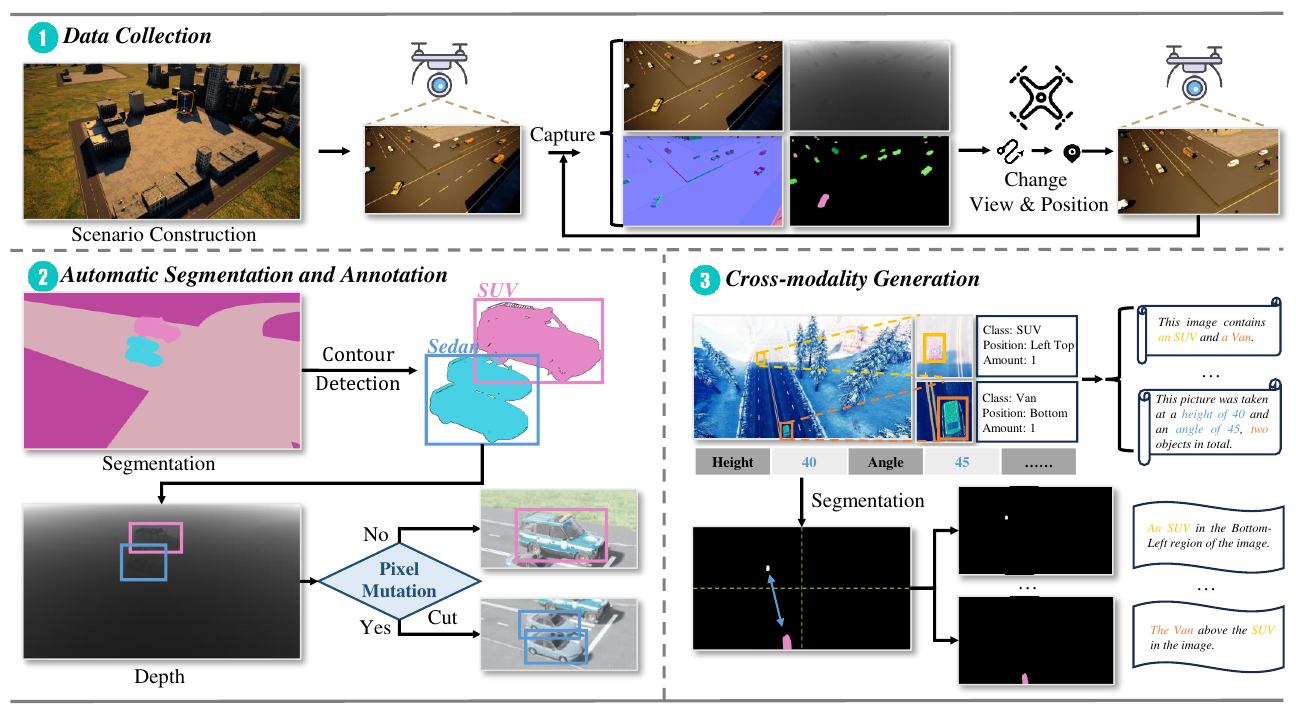}
  \caption{Pipeline of our data construction. \textbf{Step 1:} We design the automatic flight logic of the UAV to collect images from different altitudes, perspectives, and modalities. \textbf{Step 2:} We perform contour detection on the segmentation image to obtain object bounding boxes. To alleviate visually overlapped situations, we introduce the depth information, where a significant change in depth typically indicates multiple objects. \textbf{Step 3:} We extract the objects' categories, quantities, and spatial relationships to generate captions for image-text contrastive learning and referring image segmentation.
  % However, when objects of the same class are visually overlapped, their contours in the segmentation image may be merged.  utilizing the obtained segmentation and detection labels
  %Firstly, we perform contour detection on the segmentation image to obtain object bounding boxes. However, when objects of the same class are visually overlapped, their contours in the segmentation image may be merged. Therefore, we introduce the depth information, where a significant change in depth typically indicates multiple objects.
  % Then we can effectively distinguish overlapped objects and obtain more accurate annotations.
  }
  \label{method}
\end{figure*}

\textbf{AFID}~\cite{wang2023aerial} is a publicly available dataset of aerial and fluvial images, featuring detailed semantic annotation from different camera perspectives. It focuses on utilizing semantic segmentation models to fulfill unmanned hydrologic data collection, environmental inspection, and disaster warning tasks.

\textbf{SkyScenes}~\cite{khose2023skyscenes} is a synthetic dataset of densely annotated aerial images captured from UAV perspectives, containing 33.6K images from different altitudes and pitch settings. It provides pixel-level semantic, instance, and depth annotations, and enables reproducing the same viewpoint under different conditions.

\textbf{SynDrone}~\cite{10350525} proposes a multi-modal synthetic benchmark dataset containing both images and 3D data taken at multiple flying heights. It includes 28 classes of pixel-level labeling and object-level annotations for semantic segmentation and object detection.

% \begin{table}[t]
% \caption{Pixel-level UAV datasets.}
% \begin{tabular}{lcccccccc}
% \toprule
% Dataset    & Year    & S/R    & MM    & \#Modalities    & \#Images    & \#Classes    & Size [px]    & View Angle  \\
% \hline
% Stanford-Drone    & 2016    & R    & \XSolidBrush    & -    & -    & 7    & 1450×1080    & 90 \\
% UAVDT    & 2018    & R    & \XSolidBrush    & -    & 40376    & 3    & 1080×540    & Various \\
% VisDrone    & 2018    & R    & \XSolidBrush    & -    & 8629    & 10    & 1920×1080    & Various \\
% AU-AIR    & 2020    & R    & \Checkmark    & 2    & 32823    & 8    & 1920×1080    & 45 to 90 \\
% Drone-Vehicle    & 2022    & R    & \Checkmark    & 2    & 28k    & 5    & 640×512    & 90 \\
% SynDrone    & 2023    & S    & \Checkmark    & 3    & 60k    & 9    & 1920×1080    & 30,60,90 \\
% \textbf{UEMM-Air (Ours)}    & 2024    & S    & \Checkmark    & \textbf{5}    & 20k    & \textbf{13}    & 1920×1080    & Various \\
% \bottomrule
% \end{tabular}
% \label{tab:fine-grained}
% \end{table}

\section{UEMM-Air} \label{UEMM-Air}

\subsection{Scene Construction and Flight Control Logic} \label{fly}

Previous UAV-based datasets are limited in scene diversity, which tends to affect model generalization. Therefore, we aim to construct a dataset with richer scenes to improve the performance of the models. To be specific, we utilize Unreal Engine~\cite{unrealengine} with CityBLD~\cite{CityBLD2024} plugin. It can create cities of almost any size and style in a very short time to simulate scenarios in the real world. We build several scenes including cities, parks, highways, etc. We also collect a total of 13 categories and more than hundreds of vehicle models. 

We leverage Unreal Engine's movement animation to simulate dynamic scenes in reality. Employing the traffic features of Unreal Engine, we can flexibly design and construct various complex road layouts, including city streets, highways, and country roads. These layouts can precisely simulate real-world terrain and traffic conditions, providing realistic infrastructure for game scenes. Additionally, we can generate a wide variety of vehicles in the virtual environment. These vehicles can automatically navigate the generated roads based on predefined traffic rules and behaviors. By setting paths and control parameters, the vehicles can simulate real traffic flow, obey traffic signals, avoid pedestrians, and respond to traffic congestion, thereby creating highly realistic dynamic traffic scenarios.

To collect data, we control the UAV to fly and take pictures in Unreal Engine, as represented in Fig.~\ref{method} Step 1. Specifically, we build an Unmanned Aerial Vehicle simulator using AirSim and Pygame. When flying to a satisfactory shooting point, we control the UAV to fly within a height range of 5 meters to 50 meters above the horizontal surface, taking a set of photos every 5 meters up. The camera rotates from 0 degrees to 90 degrees by 5 degrees for each step. To obtain aligned pictures of different modalities, our simulator temporarily stops running when taking photos.

\begin{figure}[t]
  \centering
  \includegraphics[width=0.99\linewidth]{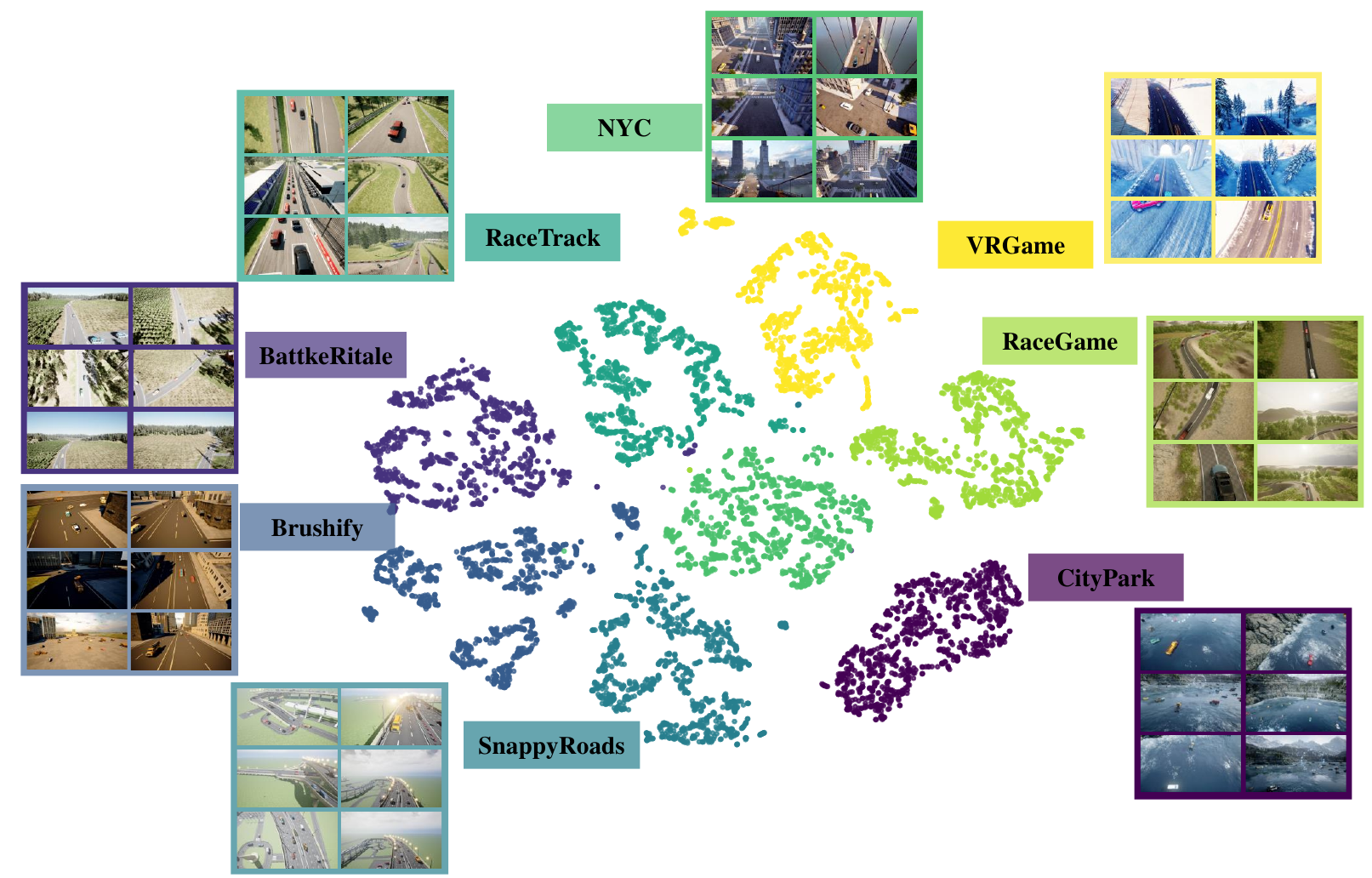}
  \caption{We randomly sampled images from various scenes and visualized the features extracted by the CLIP image encoder through T-SNE. The significant differences in features across different scenes indicate that our dataset is beneficial for enhancing the model's domain generalization performance.}
  \label{tsne}
\end{figure}

Ultimately, we built 8 different maps and collected a total of $120k$ pairs of data. Each map has its own unique style, incorporating various elements such as different lighting, scenes, and weather conditions. As shown in Fig.~\ref{tsne}, we randomly select 1,000 images from each map and utilize CLIP to extract image features for T-SNE visualization analysis. The results indicate considerable feature differences among different maps, providing rich domain representations for model training.

\subsection{Acquisition Setup} \label{setup1}

% We adopted a camera sensor setup to effectively capitalize on the AirSim simulator's capabilities while ensuring data diversity. Our acquisition pipeline equips the UAV with several co-registered sensors:
We adopt our camera sensor setup for the AirSim simulator to ensure diversity in data. 
The acquisition pipeline equips the UAV with several co-registered sensors. With the help of these sensors, we collect 5 modalities: RGB, infrared, segmentation, surface normal, and IMU parameters.

% \sethlcolor{yellow}\hl{\textit{RGB Camera}}: RGB camera offers a resolution of $1920 \times 1080$ pixels. Its vertical field of view (FoV) adjusts incrementally by $5^{\circ}$ within a range from $0^{\circ}$, indicative of the horizontal viewing angle, to $90^{\circ}$, denoting the vertical top-down viewing angle. All RGB images are stored in PNG format.

\sethlcolor{yellow}\hl{\textit{RGB Camera}}: It offers a resolution of $1920 \times 1080$ pixels. The vertical field of view (FoV) increases dynamically from $0^{\circ}$ to $90^{\circ}$, indicating that the viewing angle changes from a horizontal to a top-down view. All RGB images are stored in PNG format.

% RGB images contain rich color information, texture details, and spatial layout. They can be used to improve the contrast and clarity of images for better analysis and identification of details. Multiple overlapping aerial images can also be stitched into a large-scale, high-resolution image for further research.
RGB images contain rich color, and spatial information, facilitating better image understanding and object recognition. 
The visual image is the most common modality in computer vision tasks. However, in complex environments such as nighttime, visual images alone may not perform well due to the poor visibility and the resulting inability to effectively detect objects.

%In addition, it is an important task to quickly and accurately locate and identify objects from large-scale, high-resolution RGB aerial images.

% \sethlcolor{yellow}\hl{\textit{Depth Camera}}: Depth Camera shares the same FoV, resolution and storage format as the RGB Camera. Depth Camera interpolates each pixel value from black to white based on the depth in meters from the camera plane. The pixels with pure white means depth of 100m or more while pure black means depth of 0 meters.

\sethlcolor{yellow}\hl{\textit{Depth Camera}}: The depth camera has the same FoV, resolution and storage format as the RGB camera. 
% It interpolates each pixel value from black to white based on the depth of the distance from the camera plane. 
It interpolates each pixel value from 0 to 255 according to the depth of the distance from the camera plane. 
The white pixels show a depth of more than 100 meters, while the black pixels indicate a depth of 0 meters.

% The depth image uses pixels to represent the distance from the object in the scene to the camera. It also reflects the space shape and structure of the scene being shot. The depth images can be used to infer the height, concavity, and relative position of the object, to identify and distinguish different objects. It can also use multi-frame depth data to reconstruct the three-dimensional model of the scene for path planning and obstacle avoidance.

Depth images leverage pixels to represent the distance from the object to the camera, reflecting the spatial shape and structure of the photographed scene. Therefore, we utilize it to address the issue of inaccurate annotations caused by overlapped visual information of the objects during the annotation generation process.
It can also be leveraged to deduce an object's height, convexity, and relative position, which aids in multi-modal object objection.

% \sethlcolor{yellow}\hl{\textit{Segmentation Camera}}: Segmentation camera maintains the same FoV, resolution, and data format as the preceding cameras. Segmentation camera assigns values between 0 and 255 to objects within the Unreal Engine environment, thus ensuring accurate ground truth segmentation of scenes.

\sethlcolor{yellow}\hl{\textit{Segmentation Camera}}: The segmentation camera maintains the same FoV, resolution, and data format as the preceding cameras. It generates distinct colors for pixels belonging to different categories of objects to ensure accurate segmentation of the scene.

% The segmentation camera divides the image into multiple regions with similar attributes. The segmentation image has pixel-level information, such as each pixel is assigned to a precise category label. Using the information of the segmentation images, we can perform semantic segmentation and instance segmentation.  Using the above information, UAV can understand complex environments and perform comprehensive perception of the environment.
The segmentation image divides the image into multiple regions with similar attributes, providing pixel-level information where each pixel is assigned to a precise category label. 
%This allows for semantic segmentation and instance segmentation, achieving a comprehensive perception of the surroundings. 
Because of the detailed segmentation information, this modality can assist in the automatic generation of detection annotations. Additionally, since segmentation images inherently contain positional information, combining them with other modalities for detection often leads to improved accuracy. 
%Detailed segmentation information enhances the ability of UAVs to navigate and interact effectively in various scenarios.

% \sethlcolor{yellow}\hl{\textit{Surface Normal Camera}}: The Surface Normal Camera maps the X, Y, and Z components of the surface normal to an RGB range from 0 to 255. The contrast of the normal camera images is set as 1.5 to more distinctly delineate changes in the direction of the normals. This camera also shares the same FoV and resolution as the Scene Camera, with images saved in PNG format.

\begin{figure*}[t]
  \centering
  \includegraphics[width=0.99\linewidth]{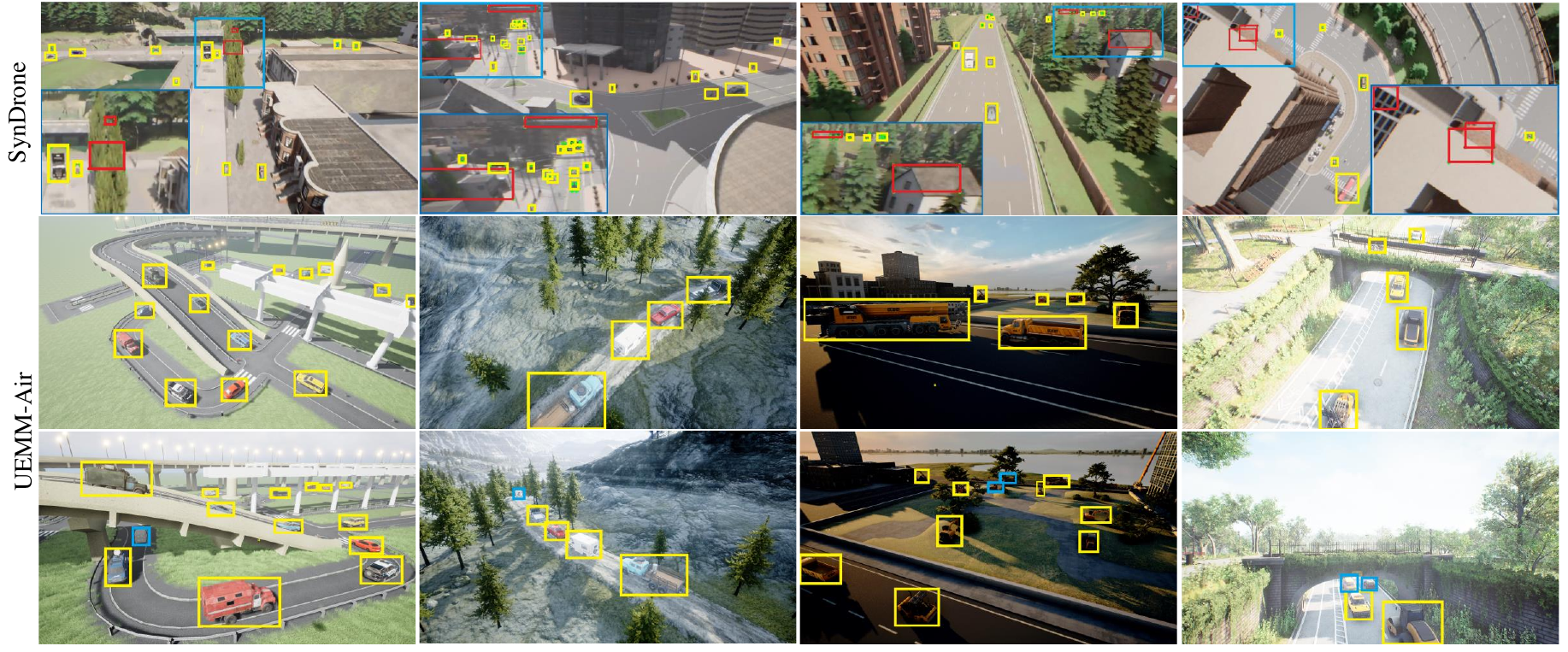}
  % \fbox{\rule[-.5cm]{0cm}{4cm} \rule[-.5cm]{4cm}{0cm}}
  \caption{Comparison of SynDrone and our UEMM-Air. Red and yellow bounding boxes indicate incorrect and correct labels, respectively. We provide two viewpoints from one scene in UEMM-Air, where blue boxes indicate originally blocked objects in the other viewpoint.
  SynDrone has incorrect labels where objects are visibly blocked, while UEMM-Air consistently demonstrates superior labeling accuracy, especially in challenging scenarios where objects are partially obscured.}
  \label{fig3}
\end{figure*}

\sethlcolor{yellow}\hl{\textit{Surface Normal Camera}}: The Surface Normal Camera maps the X, Y, and Z components of the surface normal to an RGB range from 0 to 255.  
% \red{To more clearly distinguish changes in the direction of the normals, the contrast of the normal camera pictures is adjusted to 150\%.} 
Due to the gradual changes in normal direction, it is difficult to distinguish. Therefore, the contrast of the normal camera images is set as 1.5 to more distinctly delineate changes in the direction of the normals.
This camera saves its pictures in PNG format and has the same field of view and resolution as the Scene Camera.

% Surface normal images extract geometric features of the object's surface. In object detection, these features can be used to classify and identify objects and improve the accuracy of the algorithm. Surface normal images can also reveal surface details. These surface features can generate high-precision surface models in 3D reconstruction.
Surface normal images primarily capture the geometric features and surface details of the target object. When fused with RGB images or other modalities, they can compensate for deficiencies in texture features. For example, in fine-grained object detection, the texture features introduced by the surface normal modality can help the model learn deeper fine-grained information. Additionally, it reveals intricate surface details essential for generating precise 3D reconstruction surface models.
%Surface normal images capture geometric characteristics of the object's surface. When utilized in object detection, these characteristics can aid in object classification and identification, thereby enhancing algorithm accuracy. Additionally, it can unveil intricate surface details, which are instrumental in creating precise surface models for 3D reconstruction purposes.

% \sethlcolor{yellow}\hl{\textit{IMU Parameters}}:
%  IMU Parameters include dynamic state information, GPS information, and timestamps. Dynamic state parameters include attitude angle, linear velocity, body angular velocity, linear acceleration and collective angular acceleration.
\sethlcolor{yellow}\hl{\textit{IMU Parameters}}: IMU parameters encompass dynamic state data, GPS information, flying altitude and timestamps. The dynamic state parameters consist of attitude angle, linear velocity, body angular velocity, linear acceleration, and collective angular acceleration.

% IMU parameters contains real-time attitude information and position coordinates of the UAV. The attitude information can ensure that UAV maintains a stable attitude during object detection to obtain clear images. The position coordinates can help calculate the precise position of the detection object.

IMU parameters comprise real-time attitude details and UAV position coordinates. In multimodal tasks, the current flying altitude of the UAV can be utilized to assist in determining the scale information of the object. For example, the current frame's flight posture is beneficial for the model to predict the next frame's object location, especially in tasks like video object detection or object tracking. Additionally, the UAV's GPS information can also be employed for post-detection localization tasks.
%The attitude data ensures the UAV maintains a stable orientation during object detection for capturing clear images, while the position coordinates aid in accurately determining the location of the detected object.

% \begin{table}[h]
% \centering
% \caption{Effectiveness analysis on heuristic valuation function in automatic annotation algorithm.}
% \label{ablation1}
% \begin{tabular}{ccc|c}
% \toprule
% Color & Gradient & LBP & Time [min] \\
% \hline
% \checkmark     &          &     & 497  \\
%       & \checkmark        &     & $\infty$    \\
%       &          & \checkmark   & $\infty$   \\
% \hline
% \checkmark     & \checkmark        &     & 301  \\
% \checkmark     &          & \checkmark   & 345  \\
%       & \checkmark        & \checkmark  & $\infty$   \\
% \hline
% \checkmark    & \checkmark        & \checkmark   & \textbf{282}  \\
% \bottomrule
% \end{tabular}
% \end{table}

\subsection{Automatic Image Segmentation and Annotation} \label{auto}
Most of the existing UAV-based datasets are manually annotated. Manual image annotation faces challenges in terms of accuracy and efficiency, especially when dealing with a large number of labels or low-resolution images. 

To avoid manual annotation, the SynDrone~\cite{rizzoli2023syndrone} dataset employs an automatic image labeling algorithm. They derive the absolute coordinates of UAV and vehicles from Unreal Engine, then obtain the bounding boxes of the objects by analyzing their relative position. However, this strategy causes some incorrect annotations where objects are visibly blocked but their coordinates are still marked on the image, as illustrated in Fig.~\ref{fig3}. 

In order to alleviate the problem of mislabeling in the SynDrone dataset, we propose a heuristic automatic image annotation algorithm. It makes full use of semantic and distance information from segmentation and depth images to avoid labeling visually blocked objects and mislabeling overlapped ones. Our approach is illustrated in Fig.~\ref{method}.

Employing the AirSim simulator, we assign the same color label to the same class of objects in the Unreal Engine environment.
% \red{To solve the problem of image annotation, We only focus on effectively matching model contours within segmentation images and calculating their bounding boxes.}
For each class, we convert contour detection on objects into bounding boxes and get the initial annotation. However, this step cannot recognize objects of the same category that are overlapped in the segmentation image and will mark them as one object.

To avoid mislabeling visually overlapped objects, we utilize depth images where pixel value represents the distance from the object to the camera plane to perform a secondary annotation. Intuitively, depth values mildly change on each object and a depth value jump indicates multiple objects existences.
% If there is a jump in depth, it indicates that multiple objects exist.
% means object changes too, so that the overlapping object  can be correctly segmented.
Therefore, overlapped objects can be correctly identified through depth observation.
We detect depth mutations within segmented bounding boxes to confirm object edges, adjusting labels accordingly.
% leverage the depth images to determine whether the pixel have mutated, so as to confirm the edge and perform secondary detection. 

\begin{figure*}[t]
  \centering
  \includegraphics[width=0.99\linewidth]{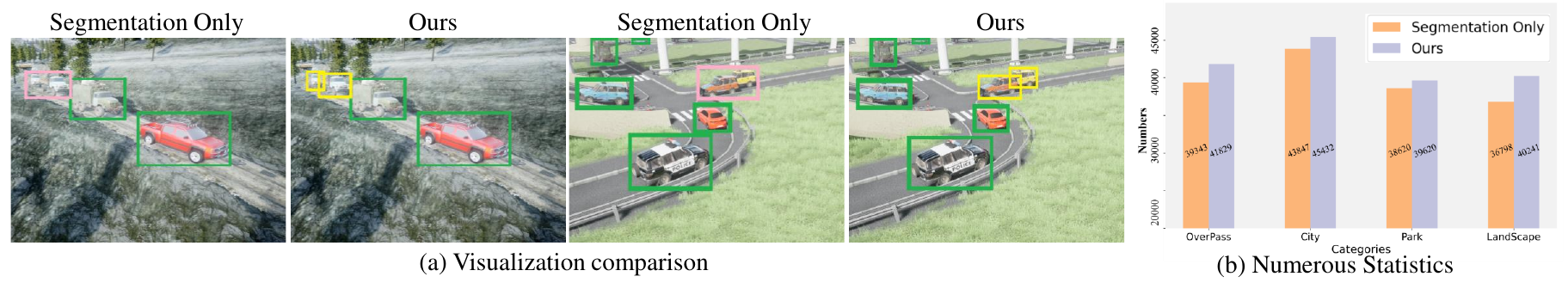}
  \vspace{-0.15cm}
  \caption{Visual (a) and Numerical (b) Comparison with only utilizing segmentation information. The pink box indicates visually overlapped objects and the yellow box shows the corrected results. Owing to overlap rectification, our approach can generate more accurate annotations.}
  %\vspace{-0.6cm}
  \label{fig4}
\end{figure*}

Fig.~\ref{fig4} presents sample annotation results for comparing our proposed algorithm with only utilizing segmentation information. It can be observed from Fig.~\ref{fig4} (a) that leveraging segmentation information alone can't effectively handle cases of visual overlapped (as shown by the pink box). Our approach can alleviate this issue by correctly distinguishing the two vehicles within the pink box. Fig.~\ref{fig4} (b) shows the numerical statistics of the annotations generated by the two methods. It can be observed that our method successfully annotates more objects, because our approach can distinguish overlapped instances and correct the annotations accurately employing depth information.
%Generally, we make full use of the information of segmentation and depth images, using a multi-source data fusion strategy to heuristically improve the accuracy of annotation. 
% Our algorithm uses color matching and contour analysis to determine object boundaries, uses depth images to distinguish overlapping objects, and detects object edges through pixel value changes. 

\subsection{Cross-modality Generation} \label{Cross}
After collecting the 5 modalities mentioned in Section~\ref{setup1}, we also need to generate the sixth modality: text. In this paper, we categorize text into two types: global and local, corresponding to the tasks of image-text contrastive learning and referring image segmentation, respectively.

\subsubsection{Image-text Contrastive Learning}
In addition to visual tasks like object detection and instance segmentation, we also hope our UEMM-Air can support vision language tasks to make UAV-based models perform zero-shot capability. In Section~\ref{auto}, we already generate accurate annotations with object bounding boxes and fine-grained class names. However, such annotations cannot be directly utilized on the text encoder of CLIP. 
Therefore, we follow the B2C method from RemoteCLIP~\cite{liu2024remoteclip} to transfer the bboxes into a set of natural language captions. However, we notice that the captions generated by RemoteCLIP only include the quantity and position of the objects relative to the image, leading to insufficient semantic information in the captions. 
To alleviate this issue, we propose a new generation approach, as shown in Fig.~\ref{method}. Specifically, we combine the scene and UAV parameter annotations provided by UEMM-Air to generate text descriptions that are more relevant to the UAV context. In addition, we provide more precise location information. Instead of being limited to the center and edges of the image, we utilize multiple sentences to comprehensively describe the distribution of the objects.

\begin{figure}[t]
  \centering
  \includegraphics[width=0.99\linewidth]{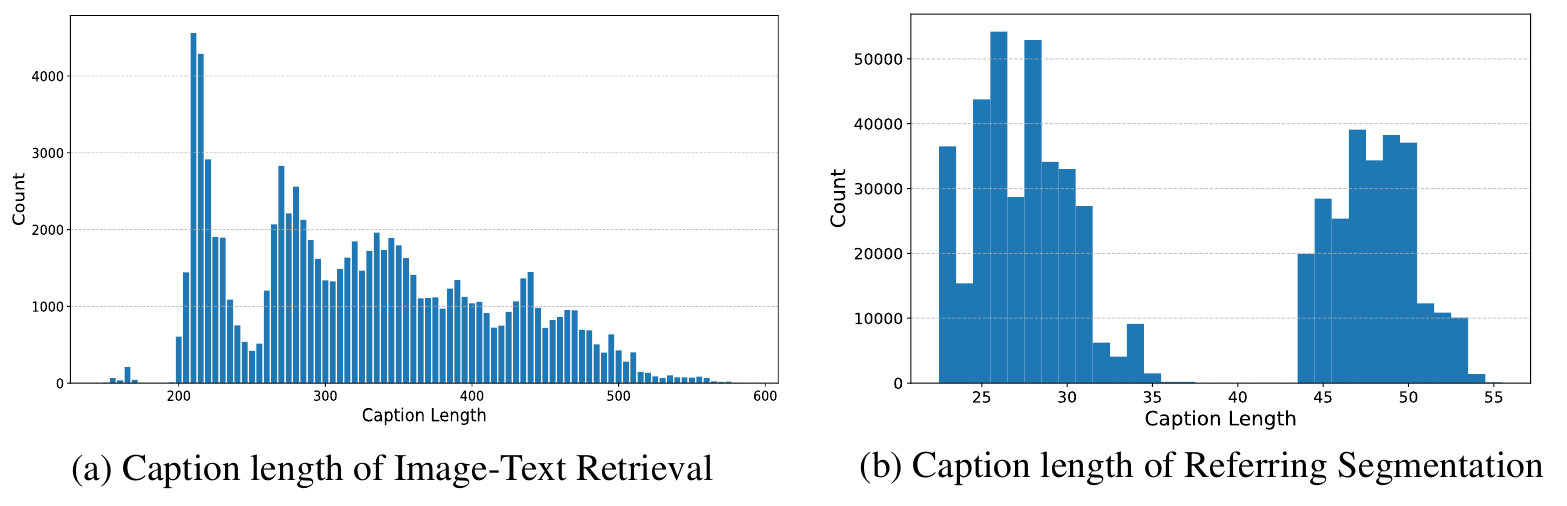}
  \caption{Distribution of captions length in our UEMM-Air.}
  \label{length}
\end{figure}

\begin{figure}[t]
  \centering
  \includegraphics[width=0.99\linewidth]{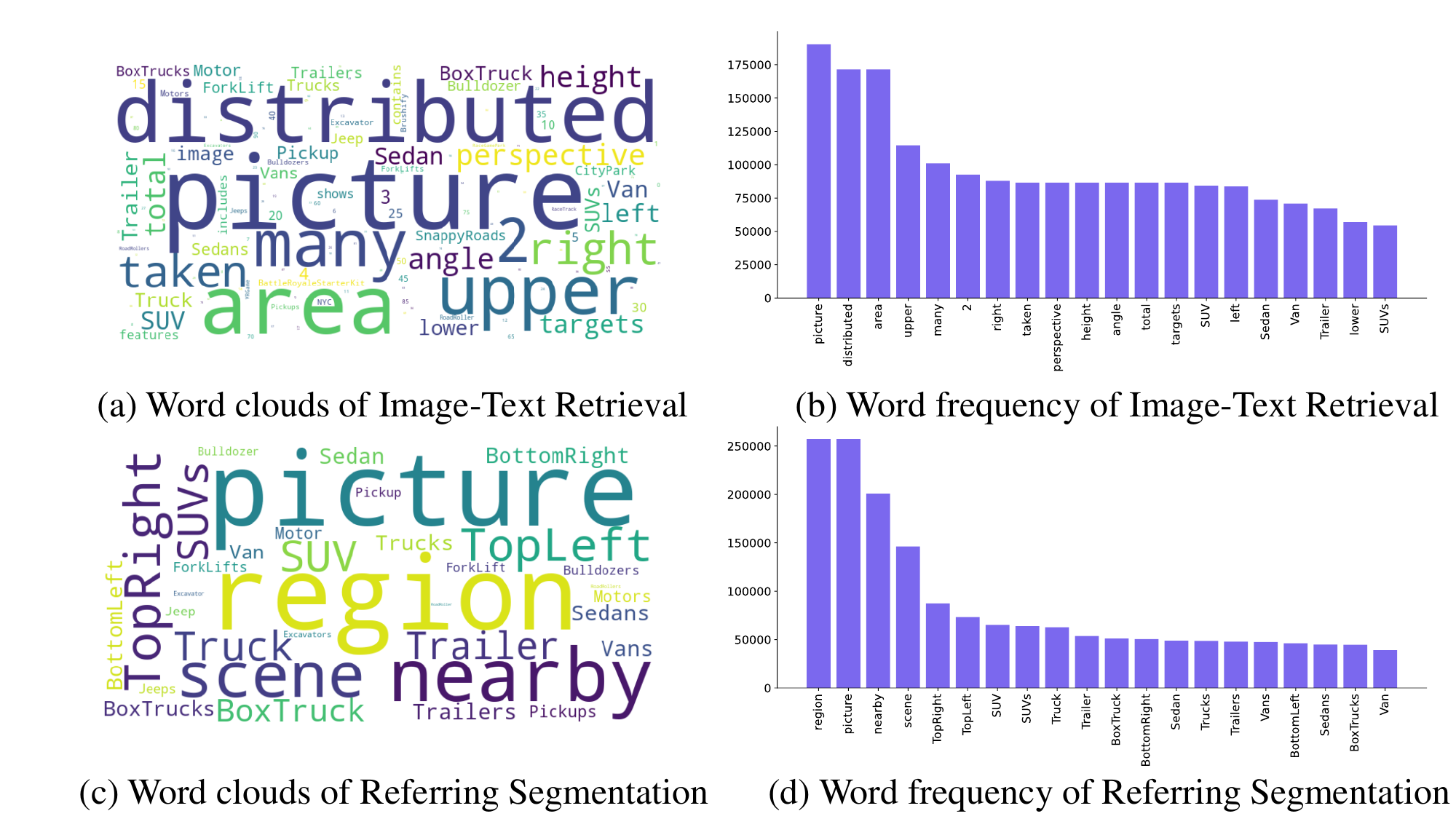}
  \caption{Word clouds and top 20 keywords of captions in our Dataset.}
  \label{wordcloud}
\end{figure}

Ultimately, we generated 7 distinct captions for each image, resulting in a total of 840,000 descriptions. We present a visualization of the caption length distribution for our final data, as represented in Fig.~\ref{length}. It can be observed that the caption length distribution shows a peak around 200, with most lengths concentrated between 200-400. Beyond 400, the frequency decreases, forming a long-tail pattern. In Fig.~\ref{wordcloud}, we also provide visualizations of word clouds and the top 20 keywords of our UEMM-Air. The words exclude stop words like "the", "is", and others.
% caption长度

\subsubsection{Referring Image Segmentation}
%Referring Segmentation
As a prominent visual-language task, referring image segmentation enables the delineation of specific objects in the visual field through natural language descriptions. This capability is equally essential in the domain of UAVs to facilitate cross-modal interactive tasks, thereby advancing embodied intelligence in UAVs. Therefore, similar to image-text contrastive learning, we propose a method for generating textual descriptions for the referring image segmentation task. By analyzing the spatial relationships of the objects within the obtained segmentation labels, we can automate the generation of the referring captions. 

Unlike referring image segmentation tasks in general domains, the UAV field often requires the segmentation of multiple objects rather than just one. For instance, it may involve segmenting a series of cars along a street. Consequently, we primarily generate descriptions in the following 3 categories: 

\begin{itemize}
    \item All objects of a specific class. \eg, \textit{All of the BoxTrucks in the image.}
    \item The spatial relationships of a particular object or certain objects relative to the image. \eg, \textit{Van in the bottom-left region of the image.}
    \item The spatial relationships of a specific object relative to other objects. \eg, \textit{The Van above the SUV in the image.}
\end{itemize}
Finally, we generated 600,000 image-mask-text pairs. Additionally, we present the statistics on captions length and word clouds in Fig.~\ref{length} and Fig.~\ref{wordcloud}, respectively. Unlike the data generated in image-text contrastive learning, the length of the captions for referential tasks is shorter, as it does not require a description of the global features of the image.

\subsubsection{Visualizations of Cross-modality Generation}

\begin{figure*}[t]
  \centering
  \includegraphics[width=0.99\linewidth]{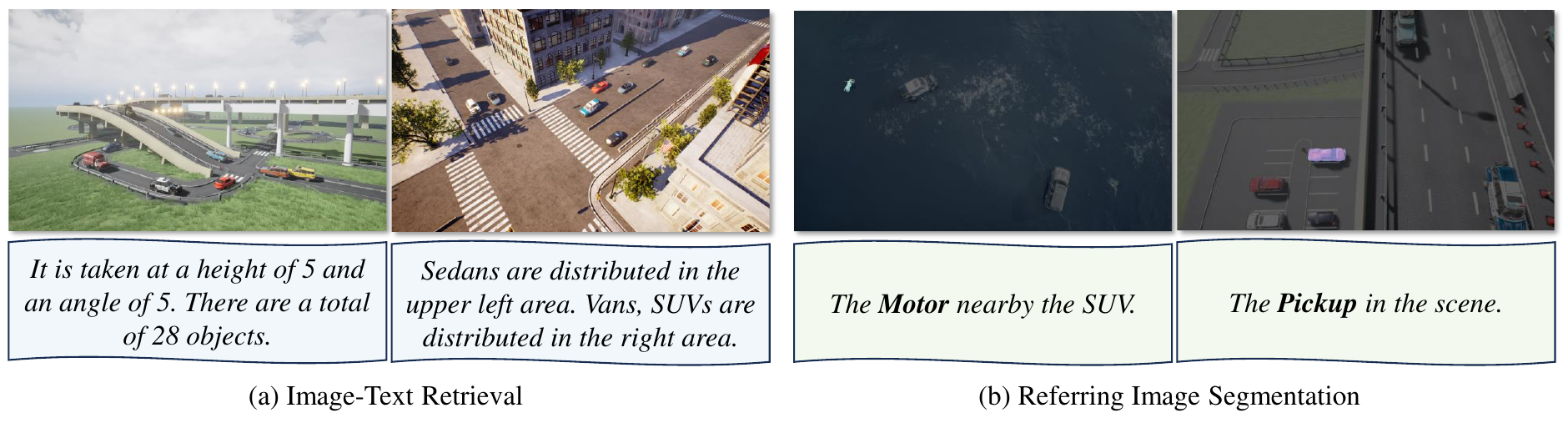}
  \caption{The visualizations for the two types of cross-modal generation. (a) In the image-text retrieval task, the text describes the global information of the image. (b) In the referring image segmentation task, the text focuses on describing the details of local objects.}
  \label{Visualizations}
\end{figure*}

To validate the accuracy of the generated captions, we randomly selected samples for visualization. As shown in Fig.~\ref{Visualizations}, the captions for the image-text retrieval task accurately describe the number and spatial relationships of the objects, as well as the UAV's height and angle information. The captions for the referring image segmentation specifically describe the location of a particular object and successfully generate the mask for the object.

\section{Benchmark and Experiments} \label{experiments}

\subsection{Experimental Setup} \label{setup}
\subsubsection{Benchmark Models}
We adopted YOLOv11~\cite{yolov11}, RT-DETR~\cite{zhao2023detrs} and D-FINE~\cite{peng2024d} as the general object detection baseline models. In the multi-modal object detection experiments, we designed a dual-path multi-modal detector with mid-level feature fusion. We utilized YOLOv7-L~\cite{wang2023yolov7} as the base detector, with two separate backbone networks to extract features from two modalities. We also designed a feature fusion module that utilizes Coordinate Attention (CA)~\cite{hou2021coordinate}. Specifically, we first directly concatenated the features of two modalities and employed Coordinate Attention to fuse them simultaneously in terms of channel and spatial information. The fused features were then entered into the neck part of the detector to complete the remaining detection tasks. 

\begin{table}[t]
\centering
%\vspace{-0.5cm}
\caption{The results of the 5-fold cross-validation experiment. Fold 1-5 represent the five randomly partitioned sub-datasets.
}
\label{5fold}
\begin{tabular}{ccccc}
\toprule
Fold & Train    & Valid & $AP_{50}$ & $AP_{75}$\\
\midrule
1    & 1,2,3,4  & 5     & 63.9\%    & 52.6\% \\
2    & 1,3,4,5  & 4     & 62.6\%    & 53.4\%\\
3    & 1,2,4,5  & 3     & 63.1\%    & 52.8\% \\
4    & 1,2,3,5  & 2     & 62.3\%    & 53.7\% \\
5    & 2,3,4,5  & 1     & 64.3\%    & 53.3\% \\
\bottomrule
\end{tabular}
\end{table}

In image-text contrastive learning experiments, we employed OpenAI CLIP~\cite{radford2021learning} and RemoteCLIP~\cite{liu2024remoteclip} as benchmark models. We selected four types of visual backbone architecture for the two CLIP models, ranging from ResNet-50, ViT-Base-16, ViT-Base-32, and ViT-Large-14. We utilized the transformer architecture, consisting of 12 layers and 8 attention heads for text encoder. The maximum token sequence length is set to 77. The InfoNCE~\cite{oord2018representation} loss operates on the [CLS] token produced by the image and text backbone.

In referring image segmentation experiments, we adopted three SoTA transformer architecture models: LAVT~\cite{yang2022lavt}, RMSIN~\cite{liu2024rotated}, RefSegformer~\cite{wu2022towards}. Among them, RMSIN is specially designed for remote sensing scenarios.

\begin{figure*}[t]
  \centering
  \includegraphics[width=0.99\linewidth]{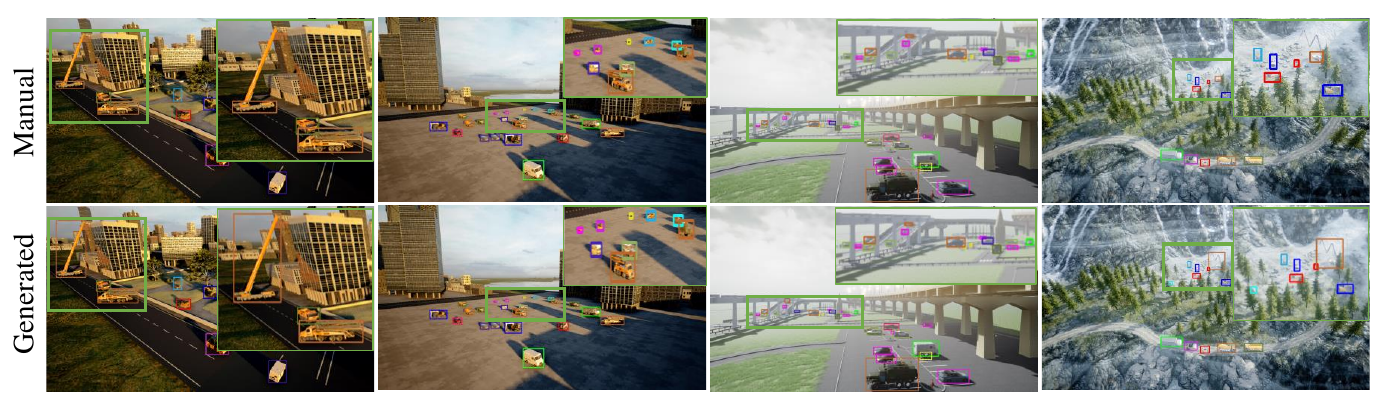}
  \caption{Comparison with Manual Annotations. Different colored boxes represent different categories. The generated annotations are basically consistent with manual annotations. Furthermore, the generated labels exhibit better fidelity compared to manual labels.}
  \label{fig5}
\end{figure*}

\begin{table}[t]
\centering
%\vspace{-0.3cm}
\caption{UEMM-Air transferability validations. We selected SynDrone dataset for comparison.}
% \vspace{-0.3cm}
\renewcommand\arraystretch{1.1}
\begin{tabular}{c|c|c|cc}
\toprule
Method                     & Fine-tuned            & Pre-trained  & $AP_{50}$ & $AP_{75}$ \\
\hline
\multirow{4}{*}{YOLOv11~\cite{yolov11}}    & \multirow{2}{*}{VisDrone} & SynDrone      & 25.6\% & 16.0\%  \\
                           &                           & UEMM-Air        & \textbf{28.3\%} & \textbf{17.7\%}  \\
                           \cline{2-5} 
                           & \multirow{2}{*}{UAV-DT} & SynDrone      & 85.4\% & 55.6\%  \\
                           &                           & UEMM-Air        & \textbf{86.2\%} & \textbf{56.1\%}  \\
\hline
\multirow{4}{*}{FasterRCNN~\cite{FasterRCNN}} & \multirow{2}{*}{VisDrone} & SynDrone      & 5.1\%  & 1.1\%   \\
                           &                           & UEMM-Air        & \textbf{5.5\%}  & \textbf{2.3\%}  \\
                           \cline{2-5} 
                           & \multirow{2}{*}{UAV-DT} & SynDrone      &   48.0\% & 9.5\%  \\
                           &                           & UEMM-Air        &  \textbf{53.8\%}       &   \textbf{14.7\%}      \\
\bottomrule
\end{tabular}
\label{tab:transfer}
\end{table}

\subsubsection{Training Settings}
All detection and segmentation experiments were conducted in Pytorch with a NVIDIA RTX 3090 GPU. During the model transferability verification, we set the batch size to 16 and trained for 200 epochs. In other object detection experiments, we froze the backbone network of the detector and trained for 50 epochs with a batch size of 32. All detectors were trained using an Adam optimizer~\cite{kingma2014adam} with a momentum of 0.937. The learning rate was initialized as 0.001 with a cosine decay~\cite{loshchilovstochastic}. We fix random seed to 18 to ensure the experiment's reproducibility.

The image-text contrastive learning experiments were trained on an NVIDIA RTX 3090 GPU. The training process was accelerated by employing the Adam optimizer~\cite{kingma2014adam}. The learning rate was set to 7e-5, 4e-5, and 1e-4, respectively, for ResNet-50, ViT-Base-32, and ViT-Large-14 models, and the corresponding batch size was set to 256, 128, and 28, respectively.

The referring image Segmentation experiments were deployed in 4 $\times$ NVIDIA RTX 4090 GPUs. The initial learning rate was set to 0.0003, with a batch size of 6 per GPU and the Adam~\cite{kingma2014adam} optimizer. The training was conducted for a total of 40 epochs. The input size of images was set to 480 $\times$ 480.

\subsection{Evaluation on Automatic Annotation Algorithm}

Considering that our labeling algorithm is auto-generated, it is necessary to validate the reliability of the labels we generate. In this section, we demonstrated the effectiveness of our labels through cross-validation experiments and visual comparisons with manual labels.

% \begin{wraptable}{r}{6cm}
% \centering
% \caption{The results of the 5-fold cross-validation experiment. Fold 1-5 represent the five randomly partitioned sub-datasets.
% }
% % \renewcommand{\arraystretch}{1.05}
% \renewcommand{\ttdefault}{pcr}
% \label{5fold}
% \begin{tabular}{cccc}
% \toprule
% Fold & Train    & Valid & $mAP_{0.5}$ \\
% \midrule
% 1    & 1,2,3,4  & 5     & 63.9\% \\
% 2    & 1,3,4,5  & 4     & 62.6\% \\
% 3    & 1,2,4,5  & 3     & 63.1\% \\
% 4    & 1,2,3,5  & 2     & 62.3\% \\
% 5    & 2,3,4,5  & 1     & 64.3\% \\
% \bottomrule
% \end{tabular}
% \end{wraptable}
\begin{table*}[t]
\centering
\caption{Comparison of fine-grained object detection results with various detectors.}
\renewcommand\arraystretch{1.1}
\resizebox{\textwidth}{!}{
\begin{tabular}{c|cc|cccccc|ccc}
\toprule
Method                   & Version & Input Size & $mAP$    & $AP_{50}$  & $AP_{75}$ & $AP_S$   & $AP_M$   & $AP_L$ & \#Params (M). & BFLOPs  & FPS (bs=1)   \\
\hline
\multirow{5}{*}{YOLOv11~\cite{yolov11}} & n                & 640×640    & 50.8\% & 68.9\% & 54.9\%  & 22.0\%  & 80.6\%  & 94.5\%  &\textbf{2.6} &\textbf{6.3} & \textbf{263.2} \\
                         & s                & 640×640    &  56.3\%      &  73.8\%      &  61.5\%      &  29.2\%      &  84.4\%      &   96.3\%     
                         &   9.4    &  21.3      &212.8\\
                         & m                & 640×640    &  57.0\%      &   74.8\%     &  62.3\%      &  30.4\%      &   84.6\%     &   96.0\%     
                         &    20.0    &  67.7      &185.2\\
                         & l                & 640×640    &   58.6\%     & 76.0\%        & 64.6\%        &32.8\%         & 85.5\%        & 96.8\%        
                         &  25.3      & 86.6       &158.7\\
                         & x                & 640×640    &  60.0\%      & 76.8\%        & 66.1\%        & 34.7\%        & 86.5\%        & 97.3\%        
                         & 56.8       &  194.5      & 94.3\\
\hline                         
\multirow{2}{*}{RT-DETR~\cite{zhao2023detrs}} & l         & 640×640           & 38.4\%       &60.3\%        &40.3\%        &11.0\%        & 64.1\%       & 84.7\%       
& 32.0       & 103.5        &90.9\\
                         & x         & 640×640           &40.1\%        &73.7\%        &38.4\%        &15.7\%        &63.1\%        &82.4\%        
                         & 65.5       & 222.5       &56.2\\
\hline
\multirow{5}{*}{D-FINE~\cite{peng2024d}}  & n                & 640×640           & 63.5\%       & 82.5\%       & 68.7\%       & 41.9\%       & 86.3\%       & 95.2\%       & 7.1       & 3.7       & 44.1\\
                         & s             & 640×640           & 71.8\%           & 90.3\%       & 78.0\%       & 54.0\%       & 89.8\%       & 96.2\%       & 24.9       & 10.2       &38.8\\
                         & m                & 640×640           & 73.2\%       & 90.7\%       & 78.8\%       & 55.6\%       & 91.4\%       & 97.1\%       & 56.4       & 19.2       &31.4\\
                         & l                & 640×640          & 74.3\%       & 91.3\%       & 80.0\%       & 57.2\%       & 91.8\%       & 97.3\%       & 90.7       & 30.7       &23.1\\
                         & x                & 640×640           & \textbf{75.8\%}       & \textbf{92.4\%}       & \textbf{81.8\%}       & \textbf{59.5\%}       & \textbf{92.5\%}       & \textbf{97.5\%}      & 61.55       & 202.2       &18.8\\
\bottomrule
\end{tabular}
}
\label{detection}
\end{table*}
\subsubsection{Verification of the Annotations' Reliability} We randomly divided the dataset into 5 parts and conducted five-fold cross-validation experiments on the YOLOv11 model. The experimental results are demonstrated in Table~\ref{5fold}. By sequentially using different sets of 4 parts as the training set and the remaining part as the validation set, we observed that the results of the 5-fold cross-validation were quite similar. The lowest $AP_{50}$ was 62.3\%, and the highest was 64.3\%, with a range of 2.0\%. This result indicates that the annotations we generated are consistent in their distribution. 

\begin{table}[t]
\caption{Comparison of segmentation results with various methods.}
\renewcommand\arraystretch{1.1}
\resizebox{0.5\textwidth}{!}{
\begin{tabular}{c|cc|ccc}
\toprule
Method                   & Version & Input\_Size & $mAP$    & $AP_{50}$  & $AP_{75}$ \\
\hline
\multirow{5}{*}{YOLOv11} & n                & 640×640    & 31.4\% & 58.8\% & 30.2\%   \\
                         & s                & 640×640    & 34.8\% & 64.3\% & 33.9\%       \\
                         & m                & 640×640    & 37.7\% & 67.9\% & 37.1\%       \\
                         & l                & 640×640    & 38.3\%      & 68.6\%      & 37.9\%       \\
                         & x                & 640×640    & \textbf{39.1\%}      & \textbf{69.5\%}      & \textbf{38.9\%}      \\
\bottomrule
\end{tabular}
}
\label{ins}
\end{table}

\begin{table}[t]
\caption{Comparison of the training over different modalities. }
\centering
\renewcommand\arraystretch{1.2}
\begin{tabular}{c|ccc}
\toprule
Modality    & $mAP$    & $AP_{50}$ & $AP_{75}$ \\
\hline
RGB         & 50.3\%   & 68.4\% & 53.2\%  \\
\hline
RGB + Seg     & 55.6\%   & 75.3\% & 54.1\%  \\
RGB + Surface & 55.8\%  & 74.3\% & 54.4\%  \\
RGB + Depth   & 53.5\%  & 73.0\% & 54.2\%  \\
\hline
RGB + Seg + Surface + Depth & \textbf{57.3\%}  & \textbf{78.2\%} & \textbf{58.0\%}  \\
\bottomrule
\end{tabular}
\label{tab:multimodal}
\end{table}
\subsubsection{Comparison with Manual Annotations} We randomly selected a subset of images for manual annotation and then visually compared them with our automatic annotations. 
As presented in Fig.~\ref{fig5}, we visualized our generated labels and manually annotated labels separately for comparison. It can be observed that our annotations are almost identical in position to the manual labels. %Additionally, due to the presence of errors in manual annotations, our generated annotations are more closely aligned with the edges of the objects.
Moreover, our annotation algorithm has some advantages in labeling small objects. We found that when objects are far away, manual annotations may contain errors due to the smaller scale of the objects. For example, manual annotations may not be as closely aligned with the edges of the objects as our generated annotations. It will introduce more foreground information, which could impact the model's accuracy.

\subsection{Evaluation on Object Detection}

In this section, we selected several mainstream detectors and conducted experiments on object detection tasks at coarse-grained and fine-grained labels and multi-modal object detection tasks. Then we conducted analysis based on the model performance and experimental results. These experimental results will serve as the baseline results of our dataset for future research.

% \begin{table*}[t]
% \centering
% \caption{Comparison of fine-grained object detection results with various detectors.}
% \begin{tabular}{l|ccccccc}
% \toprule
% Method     & BoxTruck & Bulldozer & Excavator & ForkLift & Jeep       & Motor  & Pickup  \\
% \hline
% YOLOv8     & 61.3\%   & 62.5\%    & 42.2\%    & 51.0\%   & 72.6\%     & 75.9\% & 78.7\%  \\
% FasterRCNN & 33.9\%   & 51.5\%    & 19.4\%    & 24.4\%   & 40.6\%     & 30.7\% & 50.6\%  \\
% RT-DETR    & 69.8\%   & 58.2\%    & 55.9\%    & 61.2\%   & 23.1\%     & 79.8\% & 81.8\%  \\
% \hline
% Method     & SUV      & Trailer   & Truck     & Van      & RoadRoller & Sedan  & $mAP_{50}$ \\
% \hline
% YOLOv8     & 81.9\%   & 51.1\%    & 56.1\%    & 69.0\%   & 56.7\%     & 71.6\% & 63.9\%  \\
% FasterRCNN & 46.1\%   & 30.6\%    & 29.5\%    & 35.9\%   & 39.6\%     & 40.1\% & 36.4\%  \\
% RT-DETR    & 85.1\%   & 44.3\%    & 58.5\%    & 79.2\%   & 63.2\%     & 79.4\% & 64.6\% \\
% \bottomrule
% \end{tabular}
% \label{tab:fine-grained}
% \end{table*}

\subsubsection{Transferability Verification} To demonstrate the advantage of our UEMM-Air on model transferability, We pre-trained two detectors utilizing SynDrone and UEMM-Air, respectively. Then we subsequently fine-tuned the models on the VisDrone and UAVDT datasets. The experimental results are presented in Table~\ref{tab:transfer}. While the number of images is smaller than SynDrone (20$k$ \& 60$k$), the model pre-trained on the UEMM-Air dataset demonstrates stronger generalization performance on real-world scenario data. For example, after obtaining pre-trained weights on UEMM-Air and SynDrone datasets, we fine-tuned the YOLOv8 model on the VisDrone dataset. The model pre-trained on UEMM-Air demonstrated a 2.7\% improvement in $AP_{50}$ and a 1.7\% improvement in $AP_{75}$.
This might be attributed to the provision of more accurate annotations, more categories, and more diverse scenarios in UEMM-Air for the model pre-training process.

\subsubsection{General Object Detection} 
We trained several state-of-the-art detection frameworks, including YOLOv11~\cite{yolov11}, RT-DETR~\cite{lv2023detrs}, and D-FINE~\cite{peng2024d}, utilizing our UEMM-Air. Experimental outcomes are presented in Table ~\ref{detection}. In terms of detection accuracy, the D-FINE-x model achieved the best performance with a mean Average Precision (mAP) of 75.8\%. Similar to performances on other UAV-OD datasets, the detection accuracy for small objects was 59.5\%, which is significantly lower compared to 97.5\% for large objects. This indicates that our dataset poses significant challenges for small object detection, providing valuable insights for researchers aiming to tackle the difficulties associated with small object detection. Additionally, considering the real-time requirements of UAVs, we also tested the inference time metrics. The YOLOv11-n, as a lightweight model, achieved 263.2 FPS on an RTX 3090 GPU, with only 2.6M parameters.

% \begin{table}[]
% \caption{Comparison of object detection results with various detectors.}
% \centering
% \begin{tabular}{l|cccc}
% \toprule
% Method     & LargeVehicle & MediumVehicle & SmallVehicle & $mAP_{50}$ \\
% \hline
% YOLOv8     & 55.6\%       & 73.7\%        & 75.9\%       & 68.4\%  \\
% FasterRCNN & 32.4\%       & 41.5\%        & 31.2\%       & 34.9\%  \\
% RT-DETR    & 49.8\%       & 55.8\%        &    77.5\%       &  61.1\%  \\
% \bottomrule
% \end{tabular}
% \label{tab:object-detection}
% \end{table}

\begin{table*}[t]
\caption{Cross-modal retrieval performance on UEMM-Air.}
\renewcommand\arraystretch{1.2}
\resizebox{\textwidth}{!}{
\begin{tabular}{c|c|c|c|cccc|cccc|c}
\toprule
\multirow{2}{*}{Method}     & \multirow{2}{*}{publication} & \multirow{2}{*}{Image Backbone} & \multirow{2}{*}{Text Backbone} & \multicolumn{4}{c|}{Image To Text} & \multicolumn{4}{c|}{Text To Image} & \multirow{2}{*}{Mean Recall} \\
\cline{5-12}

                            &                              &                                &                               & R@1    & R@5    & R@10   & MR     & R@1    & R@5    & R@10   & MR     &                                \\
\hline                            
\multirow{4}{*}{CLIP~\cite{radford2021learning}}       & \multirow{4}{*}{ICML2021}    & ResNet50                       & \multirow{4}{*}{Transformer}  &6.79        &27.38        & 44.05       & 26.07        & 4.57       & 19.33        & 32.41       & 18.77        & 22.42                               \\
                            &                              & ViT-B-16                       &                               & 8.67   & 30.77  & 47.35  & 28.93  & 5.96   & 22.48  & 35.71  & 21.38  & 25.15                          \\
                            &                              & ViT-B-32                       &                               & 12.80  & 43.51  & \textbf{62.72}  & \textbf{39.59}  & 8.59   & \textbf{31.09}  & \textbf{47.82}  & \textbf{29.17}  & \textbf{34.38}                         \\
                            &                              & ViT-L-14  &        &  \textbf{12.81}   &  \textbf{44.62}   &   61.99     &    39.32    &    \textbf{8.68}    &    30.68    &   46.69     &    29.08   &   34.16 \\
\hline                            
\multirow{3}{*}{RemoteCLIP~\cite{liu2024remoteclip}} & \multirow{3}{*}{TGRS2024}    & ResNet50                       & \multirow{3}{*}{Transformer}  & 7.02       &27.34        & 43.72       & 26.03       & 4.56       & 19.17       &  32.33      & 18.69       &  22.36                              \\
                            &                              & ViT-B-32                       &                               & 12.24  & 41.58  & 60.77  & 38.19  & 8.39   & 30.84  & 47.52  & 28.92  & 33.56                          \\
                            &                              & ViT-L-14 &       &  12.31  &   41.21  &    60.84    &    38.12    &   8.36     &    30.91    &   46.63     &     28.46   &    33.29        \\
\bottomrule                            
\end{tabular}
}
\label{retrieval}
\end{table*}

\begin{table*}[t]
\renewcommand\arraystretch{1.2}
\caption{Referring image segmentation performance on UEMM-Air.}
\resizebox{\textwidth}{!}{
\begin{tabular}{cc|cc|ccccc|cc}
\toprule
Methods  & Publication & Image Backbone & Text Backbone & Pr@0.5 & Pr@0.6 & Pr@0.7 & Pr@0.8 & Pr@0.9 & oIoU & mIoU \\
\hline
LAVT~\cite{yang2022lavt}     & CVPR22      & Swin-B           & BERT           & 53.17\%       & 42.42\%       & 35.47\%       & 23.69\%       &  8.03\%      &  63.72\%    & 51.09\%   \\
RMSIN~\cite{liu2024rotated}    & CVPR24     & Swin-B           & BERT           & \textbf{57.40\%}       & \textbf{49.20\%}       & \textbf{38.80\%}       & 25.40\%       &  8.60\%      &  \textbf{65.82\%}    & \textbf{51.97\%}   \\
RefSegformer~\cite{wu2022towards}     & IEEE TIP24      & Swin-B            &BERT     & 55.28\%       & 48.55\%       & 38.01\%       &  \textbf{26.04\%}      &  \textbf{8.62\%}      &  64.99\%    & 51.23\%     \\
\bottomrule
\end{tabular}
}
\label{refer}
\end{table*}
\subsubsection{Multi-modal Object Detection} 
In Table~\ref{tab:multimodal}, we conducted mid-level fusion experiments for multi-modal object detection with RGB modality and the other three modalities. The model fusion of RGB with segmentation modality achieved the best performance on $AP_{50}$, surpassing the baseline model (RGB only) by 6.9\%. The fusion of RGB with surface normal modality achieved the best performance on $AP_{75}$, surpassing the baseline model by 1.2\%. However, fusion with depth modalities resulted in the lowest performance. This could be due to the distinct features of object positions in segmentation modality and the detailed texture features in surface normal, both containing more effective information compared to depth. We also combined 4 modalities in our experiments, achieving a mAP of 57.3\%, which is an improvement of 7\% compared to utilizing the RGB modality alone.

\subsection{Evaluation on Instance Segmentation}
We selected the YOLOv11 framework to conduct instance segmentation experiments, as shown in Table~\ref{ins}. It provides benchmark results for our dataset. As the largest scale model, YOLOv11-x achieved the best accuracy, with a $mAP$ of 39.1\%, $AP_{50}$ of 69.5\%, and $AP_{75}$ of 38.9\%. Compared to object detection, instance segmentation is a more challenging dense prediction task, resulting in a relatively lower $mAP$.

% \textbf{Visualization Analyses.} 

\subsection{Evaluation on Image-Text Contrastive Learning}
Table~\ref{retrieval} presents the performance of CLIP and RemoteCLIP in image-text retrieval on our dataset. We report the retrieval recall of top-1 (R@1), top-5 (R@5), top-10 (R@10), and the mean recall of these values.
From Table~\ref{retrieval}, it can be observed that the original OpenAI CLIP performs better, achieving the highest Mean Recall of 34.38. The versions using ViT-B-32 and ViT-L-14 as visual backbones perform similarly across several metrics, but ViT-B-32 demonstrates superior average performance in both retrieval tasks. Additionally, it is noteworthy that RemoteCLIP, as the CLIP model for remote sensing, performs worse than the original CLIP. This may be due to the reduced generalization capability of RemoteCLIP after fine-tuning on remote sensing satellite data.

\subsection{Evaluation on Referring Segmentation}
We conducted experiments with three state-of-the-art referring image segmentation models on our UEMM-Air, with the results summarized in Table~\ref{refer}. The results indicate that RMSIN exhibits the best performance on both the mIoU and oIoU average performance metrics. It can be attributed to RMSIN's architecture is specifically designed for aerial perspectives, making it more conducive to feature learning from a UAV's viewpoint. Additionally, RMSIN's metrics at Pr@0.8 and Pr@0.9 are slightly lower than those of RefSegformer, aligning with the performance differences reported in the original paper for the two models. The phenomenon indirectly validates the high accuracy of the annotations generated automatically in our dataset.

\section{Conclusion} \label{discussion}
In this paper, we release a synthetic UAV-based environmental perception dataset, named UEMM-Air. 
Our work achieves three main breakthroughs:  
Firstly, to the best of our knowledge, UEMM-Air is the largest in terms of data scale, featuring the most paired modalities and the highest number of task types. Secondly, we design a new automatic annotation method, enhancing the accuracy of annotations by employing segmentation and depth images. Then, we generate a large number of text descriptions utilizing the annotations, further enriching our dataset with text modality. Finally, we provide benchmark results across multiple tasks, thereby expanding the breadth of tasks in the field of UAV-based environmental perception. We will continue to build new simulated scenarios in the future to expand the scale and number of modalities in our dataset, supporting research on UAV-based multi-modal perception tasks.
%\red{We also introduced the surface normal modality for the first time for UAV multi-modal object detection tasks.} 
% we utilized the Unreal Engine to construct various scenarios and designed a new automatic annotation method for UAV-OD tasks. 

% Finally, we conducted a series of experiments. The results validated that the models pre-trained on UEMM-Air exhibit strong transferability. We also established new benchmarks across various types of UAV-OD tasks, including fine-grained object detection and multi-modal object detection.
%Finally, we designed a new automatic annotation method for UAV-OD tasks, enhancing the accuracy of annotations by employing segmentation and depth images. 

%{\appendices
\section*{Acknowledge}
This work was supported in part by the National Natural Science Foundation of China under Grant 62372155 and Grant 62302149, in part by the Postgraduate Research and Practice Innovation Program of Jiangsu Province under Grant SJCX24\_0183, in part by the Fundamental Research Funds for the Central Universities under Grant B240201077, in part by the Aeronautical Science Fund under Grant 2022Z071108001, in part by the Qinglan Project of Jiangsu Province, and in part by Changzhou Science and Technology Bureau Project No. 20231313.

\bibliographystyle{IEEEtran}

\bibliography{NeurIPS}

\vfill

\end{document}